%% file: root.tex
\documentclass[letterpaper, 10 pt, conference]{ieeeconf}  
\let\labelindent\relax

\IEEEoverridecommandlockouts                              

\usepackage{cite}
\usepackage{amsmath,amssymb,amsfonts}
\usepackage{graphicx}
\usepackage{textcomp}
\usepackage{xcolor}
\usepackage{gensymb}
\usepackage{multirow}
\usepackage{subcaption}
\usepackage{commath} 
\usepackage{ushort} 
\usepackage{url} 

\usepackage{epsfig} 
\usepackage{balance} 

\usepackage{adjustbox}
\usepackage{booktabs, makecell, tabularx}
\usepackage{siunitx}
\setcellgapes{3.5pt}

\usepackage[ruled, vlined]{algorithm2e}
\usepackage{amssymb}
\usepackage[noend]{algpseudocode} 
\usepackage{footnote} 
\usepackage{float} 
\usepackage{kotex}
\usepackage{wrapfig}
\usepackage{setspace}
\usepackage{hyperref}
\usepackage{url}
\usepackage{setspace}
\usepackage{caption}
\usepackage{tabularx}
\usepackage{tabularray}
\usepackage{enumitem}
\usepackage{eucal}
\usepackage{bbm}

\title{\LARGE \bf
MonoDINO-DETR: Depth-Enhanced Monocular 3D Object Detection Using a Vision Foundation Model
}

\author{Jihyeok Kim$^{1}$, Seongwoo Moon$^{1}$, Sungwon Nah$^{1}$ and David Hyunchul Shim$^{1}$
\thanks{\textsuperscript{1}School of Electrical Engineering, Korea Advanced Institute of Science and Technology, Daejeon, South Korea.
        {\texttt{\{jihyeokkim, seongwoo.moon, sw.nah, hcshim\}@kaist.ac.kr}}}
\thanks{This work was partially supported by College of Engineering at KAIST, South Korea.}
}

\begin{document}

\maketitle
\thispagestyle{empty}
\pagestyle{empty}


\begin{abstract}
\input{sections/0.abstract}
\end{abstract}

\input{sections/1.introduction.tex}
\input{sections/2.relatedworks.tex}
\input{sections/3.methodology.tex}
\input{sections/4.experiments.tex}
\input{sections/5.conclusion.tex}




\bibliographystyle{IEEEtran}
\bibliography{root}

\end{document}

%% file: sections/0.abstract.tex
This paper proposes novel methods to enhance the performance of monocular 3D object detection models by leveraging the generalized feature extraction capabilities of a vision foundation model. Unlike traditional CNN-based approaches, which often suffer from inaccurate depth estimation and rely on multi-stage object detection pipelines, this study employs a Vision Transformer (ViT)-based foundation model as the backbone, which excels at capturing global features for depth estimation. It integrates a detection transformer (DETR) architecture to improve both depth estimation and object detection performance in a one-stage manner. Specifically, a hierarchical feature fusion block is introduced to extract richer visual features from the foundation model, further enhancing feature extraction capabilities. Depth estimation accuracy is further improved by incorporating a relative depth estimation model trained on large-scale data and fine-tuning it through transfer learning. Additionally, the use of queries in the transformer's decoder, which consider reference points and the dimensions of 2D bounding boxes, enhances recognition performance. The proposed model outperforms recent state-of-the-art methods, as demonstrated through quantitative and qualitative evaluations on the KITTI 3D benchmark and a custom dataset collected from high-elevation racing environments. Code is
available at \url{https://github.com/JihyeokKim/MonoDINO-DETR}.

%% file: sections/1.introduction.tex
\section{Introduction}
\label{sec:introduction}
With recent advancements in autonomous driving technology, autonomous racing competitions such as the Indy Autonomous Challenge (IAC) and the Abu Dhabi Autonomous Racing League (A2RL) that push the boundaries of innovation are gaining popularity. Among these, the IAC, the leading competition in autonomous racing, has been held at various iconic tracks such as Las Vegas Motor Speedway, Indianapolis Motor Speedway, and Monza Circuit since 2021. At CES 2025, it achieved a milestone by completing a 20-lap, 4-car autonomous race at speeds exceeding 100 MPH without any accidents.
\begin{figure}[t]
    \centering
    \begin{subfigure}{0.45\textwidth}
        \centering
        \includegraphics[width=\linewidth]{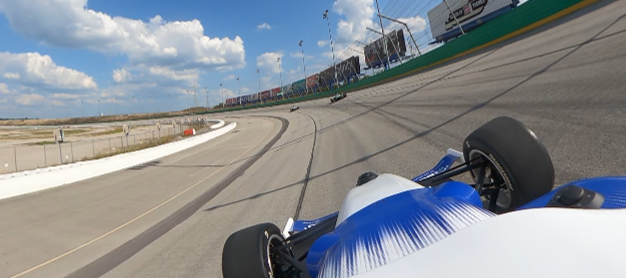}
        \caption{Illustration of a high-bank environment on a racing track, captured at Kentucky Speedway.}
        \label{fig:high-elevation}
    \end{subfigure}
    \begin{subfigure}{0.45\textwidth}
        \centering
        \includegraphics[width=\linewidth]{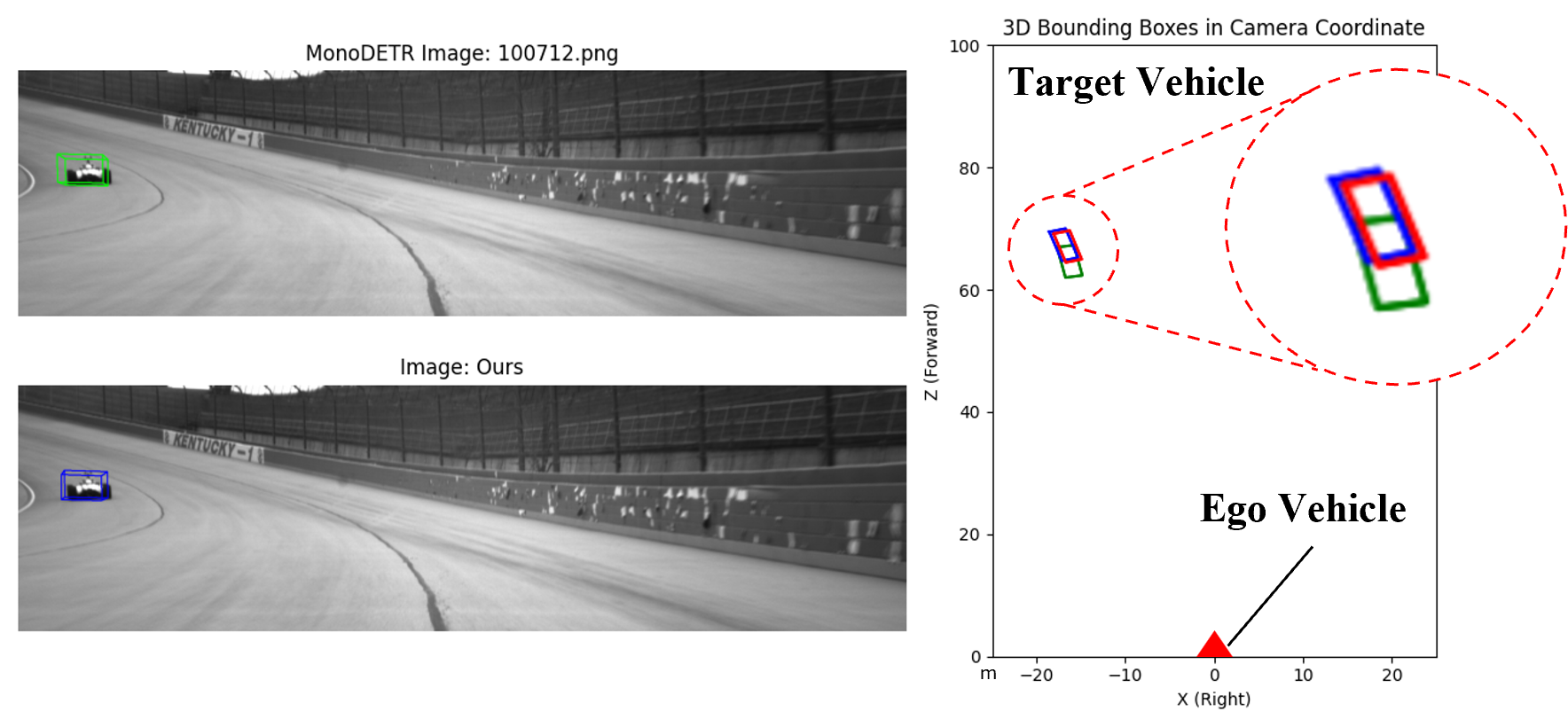}
        \caption{Inference results of MonoDETR (green), \textbf{MonoDINO-DETR} (blue), and ground truth (red) in ego-view (left) and bird's-eye view (right).}
        \label{fig:custom_result}
    \end{subfigure}
    \caption{Illustration of the racing track environment and a comparison of detection results between the proposed model and the state-of-the-art model.}
    \label{fig:main_monodinodetr}
\end{figure}

In high-speed autonomous multi-car racing, robust and reliable perception for long-range detection of the 3D position of opponent cars is crucial for overtaking. To achieve this, Indy racing cars are equipped with multiple sensors such as LiDAR, cameras, and RADAR. Although LiDAR and RADAR sensors offer significant benefits for 3D object detection tasks, they face certain limitations in racing environments. First, due to elevation changes as shown in Figure \ref{fig:high-elevation}, LiDAR points may detect not only target objects but also the ground, which can hinder the robust detection of other cars. Moreover, LiDAR points become sparser as the distance to the object increases. This sparsity can be critical for racing cars, which can reach speeds of up to 190 MPH. RADAR also has its own limitations, such as noisy data and multi-path interference, which can cause the original point to be duplicated in multiple locations even when no object is present.

Sensor fusion methods could be a solution for this problem, but the high-temperature and high-vibration conditions of racing cars make it challenging to rely on multi-sensor approaches. Since a malfunction in one sensor could lead to a critical blackout for the autonomous car, developing a robust detection system that relies on a single sensor alone would be highly beneficial in autonomous racing environments.

Cameras are currently the most attractive sensors due to their ability to extract rich features from a single input image and their relatively low cost compared to other sensors like LiDAR and RADAR. However, estimating depth from a single input image remains a challenging task because it is an ill-posed problem to infer 3D spatial positions from 2D inputs. This limitation causes Monocular 3D Object Detection (M3OD) to perform worse than LiDAR-based methods. Nevertheless, with the rapid advancements in deep learning, M3OD has shown significant improvements in recent years. Extensive research is actively being conducted in academia to further enhance its performance.

In line with these thoughts, this paper explores novel methods to enhance the performance of M3OD models in diverse environments, including urban areas with flat ground and racing tracks with significant elevation changes. By leveraging the generalized feature extraction capability of a vision foundation model and the global feature capturing ability of the Detection Transformer (DETR) \cite{carion2020end}, the proposed model demonstrates improvements in both depth and visual feature extraction, resulting in higher performance in M3OD tasks.
The main contributions of this paper are as follows:

\begin{itemize}
\item We propose \textbf{MonoDINO-DETR}, the first approach to use a foundation model, DINOv2 \cite{oquab2023dinov2}, as a backbone in M3OD, enabling the extraction of generalized features from images to improve both depth and visual feature extraction. The method is implemented as an end-to-end, one-stage detector.

\item We introduce Hierarchical Feature Fusion Block to facilitate precise object localization by leveraging multi-resolution feature information from plain Vision Transformer (ViT) \cite{dosovitskiy2020image} backbone.

\item We utilized the DETR architecture to achieve accurate depth estimation from global features without relying on additional data, such as LiDAR or depth maps. Performance is further enhanced for M3OD by incorporating 6D Dynamic Anchor Boxes.

\item The proposed method outperforms state-of-the-art models on the monocular KITTI \cite{geiger2012we} benchmark and delivers significantly improved results on a custom dataset collected in racing environments.
\end{itemize}

The remainder of the paper is organized as follows: Section \ref{sec:relatedworks} reviews related works relevant to this study. Sections \ref{sec:methodology} and \ref{sec:experiment} detail our methodology and evaluate its performance on both public and custom datasets. Section \ref{sec:conclusion} summarizes our findings and discusses future work.

%% file: sections/2.relatedworks.tex
\section{Related Works}
\label{sec:relatedworks}

\subsection{Monocular 3D Object Detection}
Monocular 3D Object Detection models aim to detect the 3D bounding boxes of target objects from a single input image. These models can be divided into three categories: 2D-detector-based, depth-image-based, and transformer-based.

\textbf{2D-Detector-Based M3OD} typically begins by localizing 2D bounding boxes and subsequently estimating 3D bounding boxes using geometric relationships or predefined 2D-3D box constraints, as demonstrated in M3D-RPN \cite{brazil2019m3d}.
MonoGround \cite{qin2022monoground} incorporates the ground plane beneath objects as a prior, transforming the ill-posed 2D-to-3D mapping problem into a more constrained and solvable task. However, these methods show poor performance due to inaccurate depth estimation and lack generalizability in diverse circumstances, as constraints like flat ground may not apply in high-elevation environments.

\textbf{Depth-Image-Based M3OD} such as D\textsuperscript{4}LCN \cite{ding2020learning} and DDMP-3D \cite{wang2021depth} first estimate depth maps from RGB images using a pre-trained depth generator while also utilizing the RGB image in a visual backbone network to extract visual features. Both types of features are extracted using Convolutional Neural Network (CNN)-based modules and fused to estimate 3D bounding boxes. 
However, as the entire network is CNN-based, it struggles to capture the global context of the image, leading to suboptimal performance. Additionally, these models often require ground-truth depth map data, which further limits their applicability.

\textbf{Transformer-Based M3OD} methods have been recently proposed, showing promising performance. MonoDTR \cite{huang2022monodtr} exploits a Transformer \cite{vaswani2017attention} encoder-decoder architecture to globally integrate context and depth-aware features, requiring LiDAR data for auxiliary supervision. In contrast, MonoDETR \cite{zhang2023monodetr} uses the DETR \cite{carion2020end} architecture to predict 3D bounding boxes and estimates foreground depth maps for supervision without relying on additional data. Both models utilize depth distributions for each pixel, following the approach introduced in CaDDN \cite{reading2021categorical} for supervision. Although these transformer-based models improve global feature extraction for depth estimation, they still face limitations due to their reliance on CNN backbones, which struggle to effectively capture global features.

\subsection{Detection Transformer}
DETR\cite{carion2020end} is an end-to-end object detector that eliminates the need for hand-crafted anchor boxes by combining a CNN backbone with a transformer architecture to model global relationships for object detection. Despite its strong performance, DETR faces challenges such as slower training convergence due to its computational complexity and inefficiency in matching object queries.
DAB-DETR\cite{liu2022dab} improves DETR's efficiency and accuracy by introducing 4D anchor-based queries that are dynamically updated during decoding, enabling effective bounding box refinement. The proposed model extends this idea by incorporating 6D dynamic anchor boxes to better handle asymmetric shapes which achieves improved performance in M3OD task.

\subsection{Vision Foundation Model}
Vision Foundation Models (VFMs) are large-scale pre-trained models designed for versatile vision tasks, such as object detection, semantic segmentation, and depth estimation, leveraging extensive datasets like ImageNet-1k \cite{russakovsky2015imagenet}. Unlike traditional CNN-based backbones, such as ResNet \cite{he2016deep} and DenseNet \cite{huang2017densely}, VFMs like CLIP \cite{radford2021learning}, DINO \cite{caron2021emerging}, SAM \cite{kirillov2023segment}, and DINOv2 \cite{oquab2023dinov2} are based on ViT \cite{dosovitskiy2020image} backbones, which provide richer contextual features for a wide range of applications. In this paper, DINOv2 is selected as the backbone for the M3OD task to enhance depth estimation and 3D object detection performance.

%% file: sections/3.methodology.tex
\begin{figure*}[h]
    \centering
    \includegraphics[width=14cm]{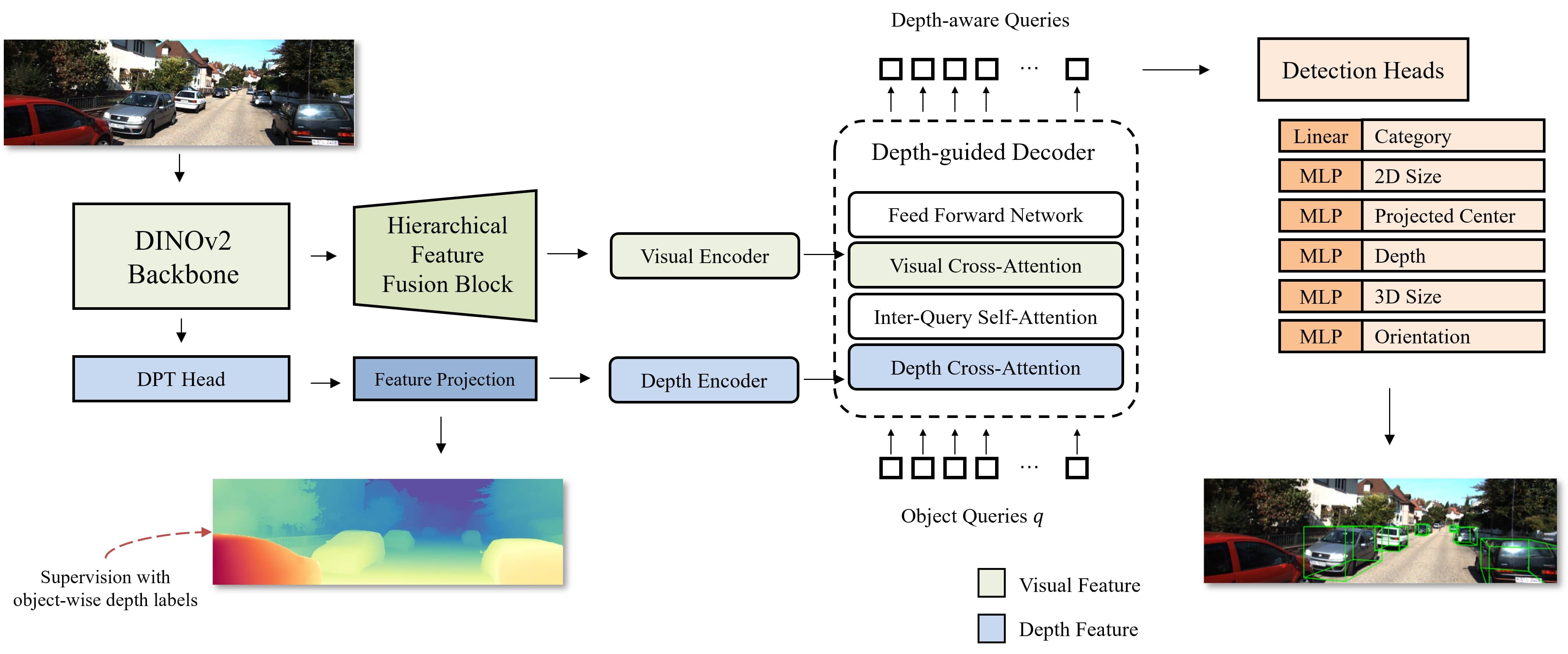}
    \caption[Overall Structure of MonoDINO-DETR]{\textbf{Overall Structure of MonoDINO-DETR.} The proposed method, \textbf{MonoDINO-DETR}, is composed of four main components: the Feature Extraction Module, the Object-Wise Supervision Module, the Depth-Aware Transformer, and the MLP-Based Detection Heads. The visual feature extraction process is represented in green, while the depth feature extraction process is represented in blue.
    } \label{fig:Monodinodetr}
\end{figure*}
\section{Methodology}
\label{sec:methodology}
The overall architecture of the proposed model is shown in Figure \ref{fig:Monodinodetr}. Given a single input image, visual features for object detection and depth features for depth estimation are extracted using the foundation model backbone with a Hierarchical Feature Fusion Block (HFFB) and the backbone with a Dense Prediction Transformer (DPT) \cite{ranftl2021vision} head, respectively. The depth features are then projected to generate a one-dimensional depth map, which is supervised using ground truth bounding box and center-depth data. Both visual and depth feature are fed into a Detection Transformer network to generate depth-aware queries. Finally, Multi-Layer Perceptron (MLP)-based detection heads estimate the class, 2D size, projected center point on the image, depth, 3D size of the bounding box, and orientation of the target objects.

\subsection{Feature Extraction Module}
\begin{figure*}[h]
    \centering
    \includegraphics[width=14cm]{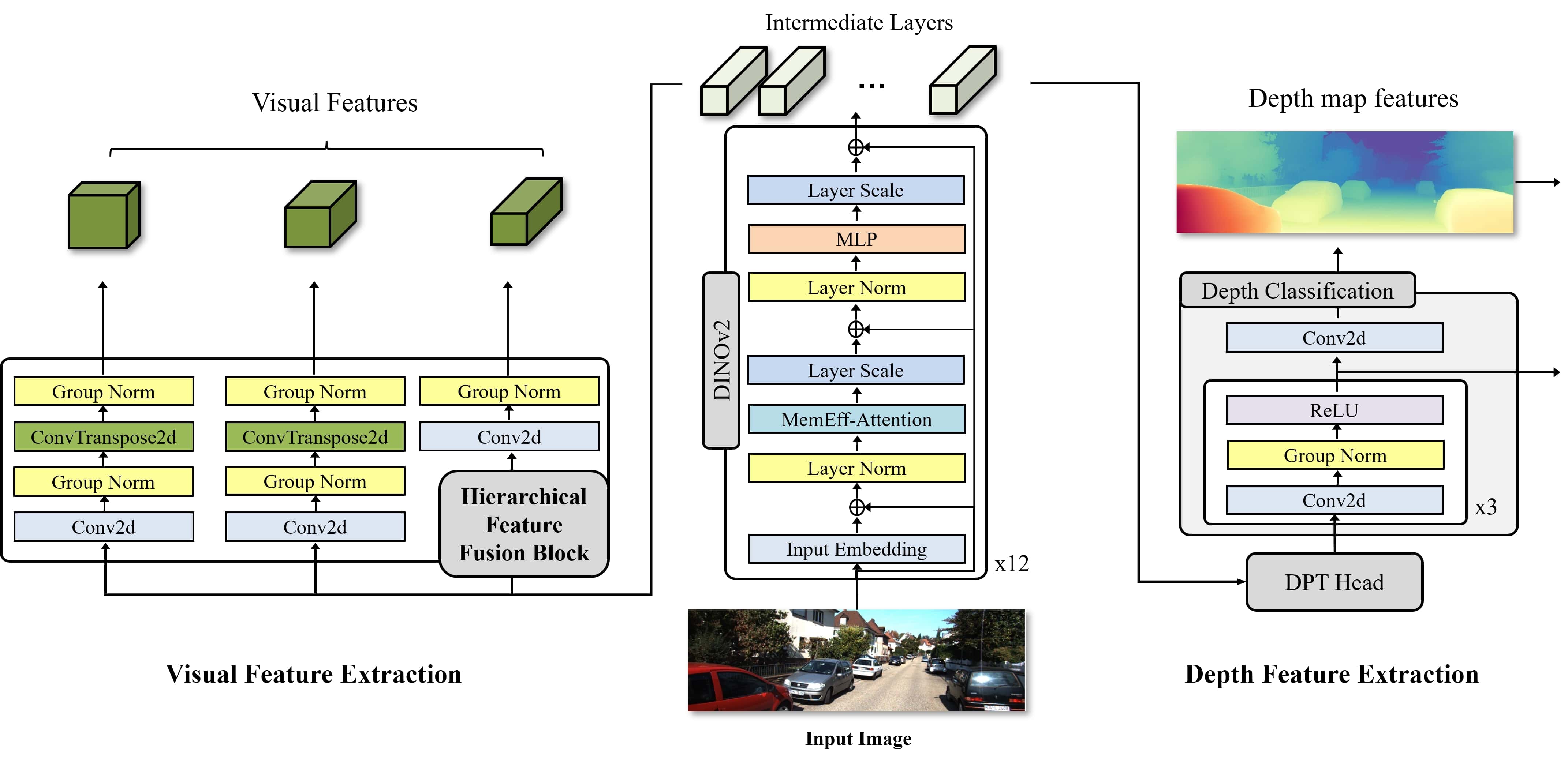}
    \caption[Overall Structure of Feature Extraction Module]{\textbf{Overall Structure of Feature Extraction Module.} The Feature Extraction Module is divided into three components: the DINOv2 backbone, the Visual Feature Extraction Module, and the Depth Feature Extraction Module. The \textbf{Hierarchical Feature Fusion Block} serves as the key module for visual features, while the combination of the DPT Head and DINOv2, which together form the \textbf{Depth Anything V2} architecture, serves as the key module for depth features.
    } \label{fig:Feature_extraction_module}
\end{figure*}

\textbf{DINOv2 Backbone.} The DINOv2 \cite{oquab2023dinov2} backbone extracts general-purpose features from the input image. Starting with the interpolated input image $I \in \Bbb{R}^{H\times W \times 3}$, the image is divided into patches of size 14x14 pixels, denoted as $x_{p} \in \Bbb{R}^{N \times (P^2 \cdot C)}$. Here, $(H, W)$ represents the resolution of the interpolated image, $C$ denotes the number of channels, $(P, P)$ corresponds to the resolution of each patch $(P=14)$, and $N=HW/P^2$ is the total number of patches, which also serves as the input sequence length for the Transformer. To preserve spatial information, positional embeddings are added to the patch embeddings and the flattened patches are projected to dimension $D=768$.
Using a memory-efficient attention module and sequential MLP modules, the model generates features that capture global context.
This process is repeated 12 times, and the features from the 3rd, 6th, 9th, and last layers are reshaped and used for the next step. These features are denoted as $f_{\frac{1}{14}}^3, f_{\frac{1}{14}}^6, f_{\frac{1}{14}}^9, f_{\frac{1}{14}}^{12}$, where $\frac{1}{14}$ represents the downsampling ratio relative to the original input size. This process is illustrated in Figure \ref{fig:Feature_extraction_module}.

\textbf{Hierarchical Feature Fusion Block.} Unlike typical CNNs, which employ multi-scale hierarchical architectures tailored for detection-specific tasks, DINOv2 features a plain, non-hierarchical design that maintains a single-scale feature map throughout. While its inclusion of global context is advantageous for depth estimation tasks, the absence of a hierarchical structure may limit its performance in object detection tasks. To address this and fully leverage its features, we construct a hierarchical architecture using transposed convolutions and group normalization to enhance visual feature extraction for object detection.

For visual feature extraction, the last three of the four intermediate features, $f_{\frac{1}{14}}^6, f_{\frac{1}{14}}^9, f_{\frac{1}{14}}^{12}$, are utilized to generate hierarchical features as illustrated in Figure \ref{fig:Feature_extraction_module}. Each feature undergoes a convolutional layer to adjust its dimensions, followed by transposed convolutions with varying kernel sizes and strides. This process produces features with resolutions that are 4, 2, and 1 times larger than the original, resulting in $f_{\frac{4}{14}}', f_{\frac{2}{14}}',$ and $f_{\frac{1}{14}}'$, respectively.
These multi-scale features enhance the object detection performance of the original backbone, which is a plain ViT model with limited hierarchical capabilities. The importance of the Hierarchical Feature Fusion Block will be demonstrated in the ablation study.

\textbf{Depth Estimation with Transfer Learning.} For depth feature extraction, the DPT \cite{ranftl2021vision} head is employed. Although not a foundation model, Depth Anything V2 \cite{yang2024depth} is a large-scale pre-trained model designed for relative depth estimation, trained on synthetic depth images using a teacher-student framework. It adopts a similar architecture, combining a DINOv2 backbone with a DPT head.
By utilizing this architecture for our depth feature extraction module, we leverage the pre-trained weights of Depth Anything V2 and apply transfer learning to estimate absolute depth values for each pixel.

In our depth feature extraction module, the intermediate features from the backbone, $f_{\frac{1}{14}}^3, f_{\frac{1}{14}}^6, f_{\frac{1}{14}}^9, f_{\frac{1}{14}}^{12}$, are projected to specific dimensions: $D = 256, 512, 1024, 1024$, respectively. Each feature is then upsampled or downsampled to achieve resolutions of 4, 2, 1, and 1/2 times the original resolution, respectively. Features from each layer are subsequently combined using a RefineNet-based feature fusion block \cite{lin2017refinenet, xian2018monocular}. Additional convolutional layers are applied to estimate absolute depth bins for each pixel, which are supervised within the object-wise supervision module. Finally, the depth map features are obtained, as shown in Figure \ref{fig:Feature_extraction_module}.

\subsection{Object-Wise Supervision Module}
To effectively incorporate depth information into the depth features, the depth map is supervised following the method in MonoDETR \cite{zhang2023monodetr}. This approach relies solely on discrete object-wise depth labels derived from the ground-truth depth information of the target objects, without requiring additional dense depth annotations.

First, $k+1$ discretized depth bins are created using Linear Increasing Discretization (LID)\cite{reading2021categorical}, and each pixel is assigned to its corresponding depth bin.
By using wider intervals for farther objects in LID, the model becomes more robust to small depth estimation errors at greater distances. Foreground object labels are then used to supervise each pixel within a bounding box by assigning them to the central depth class of the object. For supervision, the Focal Loss \cite{lin2017focal}, $\mathcal{L}_{dmap}$, is employed to balance the contributions of the background and objects:

\begin{equation}
\mathcal{L}_{dmap} = -\alpha (1-p_t)^\gamma \log(p_t)
\end{equation}

\noindent where $\alpha \in [0, 1] $  denotes the balancing factor and $\gamma \in [0, 5]$ denotes the modulating factor.

\subsection{Depth-Aware Transformer with Dynamic Anchor Boxes}
\textbf{Depth-Aware Transformer.} The visual and depth map features extracted by the feature extraction module are fed into separate encoders: a visual encoder for visual features and a depth encoder for depth features. Following the MonoDETR \cite{zhang2023monodetr} design, these encoders are paired with a shared decoder block.

\begin{figure}[h]
    \centering
    \includegraphics[width=0.9\linewidth]{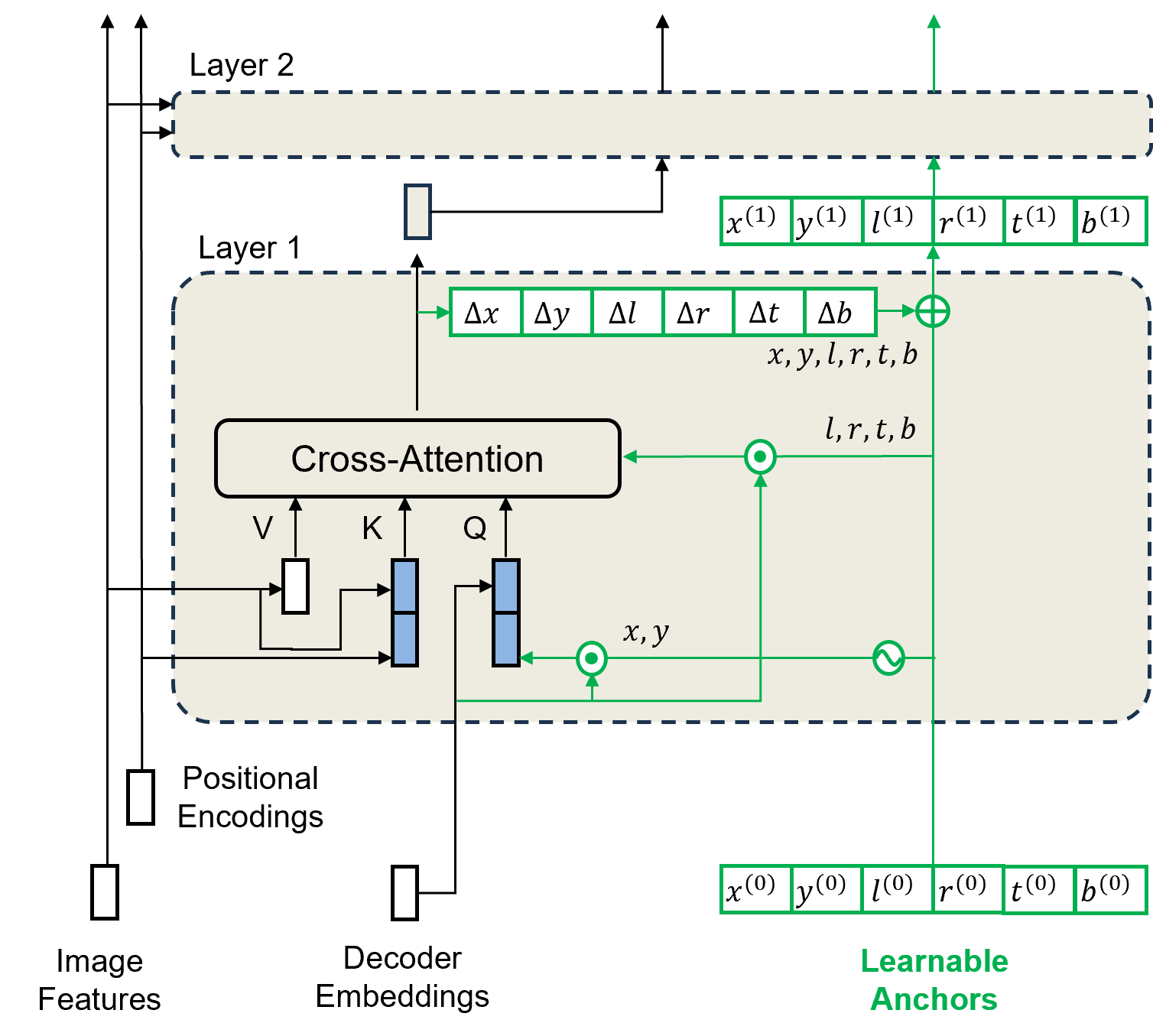}
    \caption[Representation of 6D Dynamic Anchor Boxes]{\textbf{Representation of 6D Dynamic Anchor Boxes.} 6D DAB extends 4D DAB by refining the reference point $(x, y)$ and the distances to the left, right, top, and bottom edges $(l, r, t, b)$ at each layer. This iterative refinement improves the model's adaptability to asymmetric object shapes.
    } \label{fig:6D-DAB-DETR}
\end{figure}

\textbf{Enhanced Detection with 6D Dynamic Anchor Boxes.} In the decoder architecture, Dynamic Anchor Boxes (DAB), as introduced in DAB-DETR \cite{liu2022dab} are utilized. Unlike the vanilla DETR model, DAB-DETR sets the dimension of object queries to 4, enabling the effective estimation of a reference query point $(x, y)$ and a reference anchor size $(w, h)$. These anchor boxes are dynamically updated layer by layer. In our model, DAB is extended to six dimensions to iteratively refine anchor boxes for better handling of asymmetric shapes as shown in Figure \ref{fig:6D-DAB-DETR}. The reference point $(x, y)$ and the distances from the center to the left, right, top, and bottom edges $(l, r, t, b)$ are iteratively refined at each layer, improving adaptability to complex object shapes.

\subsection{MLP-Based Detection Heads}
After passing through the decoder, the depth-aware queries are processed by a series of MLP-based heads to predict various attributes, such as the object category, 2D size, projected 3D center, depth, 3D size, and orientation.
To ensure accurate alignment between the predicted queries and ground-truth objects, the loss for each query-label pair is computed.  As in MonoDETR \cite{zhang2023monodetr}, the losses for the six attributes are categorized into 2D and 3D groups.
The Hungarian Algorithm \cite{kuhn1955hungarian} is then employed to determine the optimal matching. Since 2D attributes are typically predicted with higher accuracy compared to 3D attributes, only the Loss 2D value is used as the matching cost.
Following this matching step, $N_{gt}$ valid pairs are obtained from a total of $N$ queries, where $N_{gt}$ represents the number of ground-truth objects.
The overall loss for training is then formulated as follows:

\begin{equation}
\mathcal{L}_{overall} = \frac{1}{N_{gt}} \cdot \sum_{n=1}^{N_{gt}}{(\mathcal{L}_{2D}+\mathcal{L}_{3D})} + \mathcal{L}_{dmap}
\end{equation}

%% file: sections/4.experiments.tex
\section{Experiments}
\label{sec:experiment}
\subsection{Experimental Setup}
\textbf{Dataset.} To validate the performance of the proposed model in diverse environments, both a public dataset and a custom dataset are utilized. For the public dataset, the widely-adopted \textbf{KITTI} 3D Benchmark \cite{geiger2012we} is selected. Following standard practice \cite{chen20153d, chen2016monocular}, the dataset is splited into 3,712 samples for training and 3,769 samples for validation.

\begin{figure}[t]
    \centering
    \includegraphics[width=\linewidth]{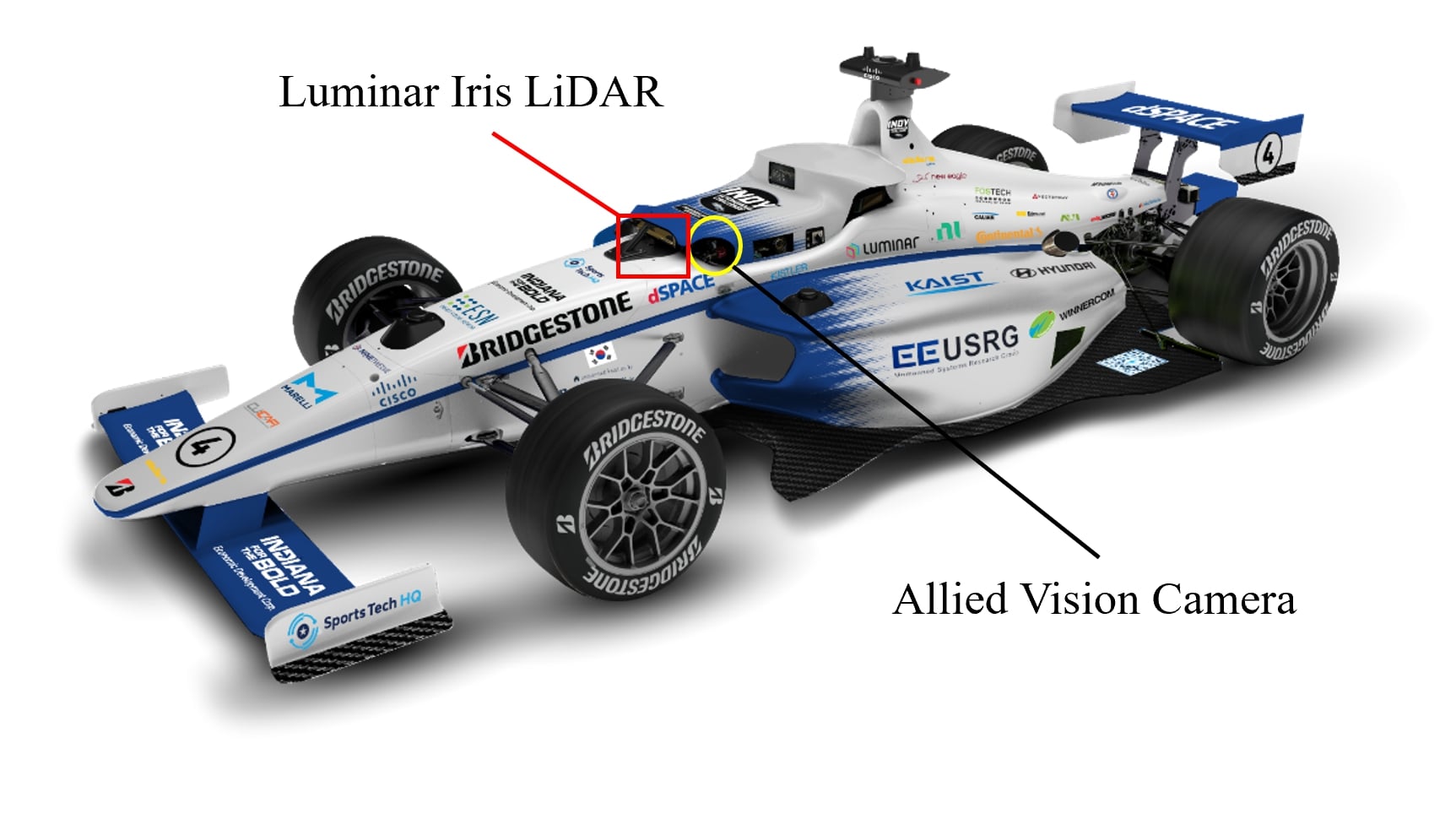}
    \caption[Indy Race Car Platform]{\textbf{Indy Race Car Platform.} Synchronized camera image data and LiDAR data were acquired during the race using a front-mounted Luminar Iris LiDAR sensor and a front-left Mako G-319C camera.
    } \label{fig:Indy car platform}
\end{figure}

Most M3OD models in academia are primarily evaluated on public datasets like KITTI, which consist of flat roads with minimal elevation changes. To assess the model's performance in a high-bank environment, such as a race track, we constructed a \textbf{custom dataset} from a racing scenario. Image and LiDAR data were collected using our race car platform, as shown in Figure \ref{fig:Indy car platform}, during a head-to-head race with an opponent vehicle at the Kentucky Speedway, where each car took turns overtaking the other in a parallel line while progressively increasing the target speed.

First, we collected time-synchronized image and LiDAR data to create paired image-LiDAR samples. Ground-truth 3D bounding boxes were generated using a pseudo-labeling approach with the PointPillars model \cite{lang2019pointpillars}, a LiDAR-based detection method. The resulting data was then post-processed and filtered to obtain 1,171 training samples and 293 validation samples, each containing paired images and 3D bounding box labels in the KITTI format.

\textbf{Evaluation metrics.}  We report the detection results for three levels of difficulty-easy, moderate, and hard-on the KITTI validation dataset.
The evaluation is conducted using the average precision of bounding boxes in 3D space, $AP_{3D}$, and the bird's-eye view, $AP_{BEV}$, both calculated at 40 recall positions.

\textbf{Implementation details.} For training, we used 4 NVIDIA TITAN RTX GPUs for 195 epochs with a batch size of 8 and a learning rate of 0.0002. The AdamW \cite{loshchilov2017decoupled} optimizer with a weight decay of 0.0001 was employed. The learning rate was reduced by a factor of 0.1 at 125 and 165. For the foreground depth map, the depth range $[d_{min}, d_{max}]$ was set to $[0m, 60m]$ for the KITTI dataset, and $[0m, 120m]$ for the custom dataset. The number of bins $k$ was set to 80 and 160, respectively.

\begin{table*}[h]
\caption[Comparison of our model with state-of-the-art models on KITTI \textit{val.}]{\textbf{Comparison of our model with state-of-the-art models on KITTI \textit{val.} set for the car class.} `Mod.' indicates the moderate difficulty level. Bold numbers highlight the best results, underlined numbers indicate the second-best results, and blue numbers represent the improvement over them. {\small *Since CaDDN uses a substantial amount of GPU memory, the batch size is set to 2 per GPU across 4 GPUs for CaDDN, and 8 for other models.}}
\centering
\small
\begin{tabular}{l|c|ccc|ccc|c}
	\toprule
\multirow{2}{*}{Method} & \multirow{2}{*}{Extra data} & \multicolumn{3}{c|}{Val,\ $AP_{3D}$} & \multicolumn{3}{c|}{Val,\ $AP_{BEV}$} & Time \\ 
& & Easy & Mod. & Hard & Easy & Mod. & Hard & (ms) \\
\midrule
CaDDN* (CVPR 2021) \cite{reading2021categorical} & \multirow{2}{*}{LiDAR} & 21.91 & 15.28 & 13.66 & 29.96 & 21.61 & 18.95  & - \\
MonoDTR (CVPR 2022) \cite{huang2022monodtr} &  & 23.92 & \underline{18.76} & \underline{15.81} & \underline{32.24} & \underline{24.66} & \underline{21.21}  & - \\
\midrule
MonoGround (CVPR 2022) \cite{qin2022monoground} & \multirow{2}{*}{None} & 19.78 & 14.46 & 12.42 & 28.11 & 21.21 & 19.00  & 42 \\
MonoDETR (ICCV 2023)\cite{zhang2023monodetr} &  & \underline{24.29} & 17.52 & 15.28 & 32.16 & 23.54 & 20.12 & 23 \\
\midrule
MonoCD (CVPR 2024) \cite{yan2024monocd} & Planes & 21.39 & 15.86 & 13.09 & 29.60 & 22.73  & 13.09 & 35 \\
\midrule
\textbf{MonoDINO-DETR} & \multirow{2}{*}{None} & 26.72 & 19.19 & 15.92 & 37.65 & \textbf{26.70} & 21.79 & 66 \\
\textbf{MonoDINO-DETR + DAB} &  & \textbf{27.93} & \textbf{19.39} & \textbf{15.97} & \textbf{38.51} & 26.15 & \textbf{22.00} & 74 \\
\textit{Improvement} & \textit{v.s. second-best} & \color{blue}{+3.64} & \color{blue}{+0.63} & \color{blue}{+0.16} & \color{blue}{+6.27} & \color{blue}{+1.49} & \color{blue}{+0.79} & \\
\bottomrule
\end{tabular}
\label{tab:KITTI_valid}
\end{table*}

\begin{table*}[h]
\caption[Comparison of our model with state-of-the-art models on our custom dataset]{Comparison of our model with state-of-the-art models on our custom dataset for the car class.}
\centering
\small
\begin{tabular}{l|c|cc|cc|c}
	\toprule
\multirow{2}{*}{Method} & \multirow{2}{*}{Extra data} & \multicolumn{2}{c|}{Val,\ (IoU = 0.7)} & \multicolumn{2}{c|}{Val,\ (IoU = 0.5)} & Time \\ 
& & $AP_{3D}$ & $AP_{BEV}$  & $AP_{3D}$ & $AP_{BEV}$ & (ms) \\
\midrule
MonoGround (CVPR 2022) \cite{qin2022monoground} & \multirow{2}{*}{None} & 1.49 & 10.14 & 21.66 & \underline{71.87} & 110 \\
MonoDETR (ICCV 2023)\cite{zhang2023monodetr} &  & \underline{9.86} & \underline{21.59} & \underline{36.35} & 44.09 & 23 \\
\midrule
\textbf{MonoDINO-DETR {\small (small)}} & \multirow{2}{*}{None} & 22.47 & 41.44 & 62.11 & 69.92 & 41 \\
\textbf{MonoDINO-DETR {\small (base)}} &  & \textbf{26.23} & \textbf{59.68} & \textbf{80.30} & \textbf{88.80} & 70 \\
\textit{Improvement} & \textit{v.s. second-best} & \color{blue}{+16.37} & \color{blue}{+38.09} & \color{blue}{+43.95} & \color{blue}{+16.93} & \\
\bottomrule
\end{tabular}
\label{tab:Cusom_valid}
\end{table*}

\subsection{Main Results}
\textbf{KITTI validation set.} Table \ref{tab:KITTI_valid} presents the validation result of state-of-the-art models and our model on the KITTI dataset. The proposed model, MonoDINO-DETR, outperforms all recent models in the table in terms of both $AP_{3D}$ and  $AP_{BEV}$. Even without requiring any additional data, it surpasses models that utilize extra data, such as LiDAR or ground information.
Furthermore, even without DAB, it still demonstrates the second-best performance. Visualized results for the KITTI dataset are shown in Figure \ref{fig:KITTI visualized}.

\textbf{Custom Dataset.} Table \ref{tab:Cusom_valid} presents the validation results for the custom dataset.
We tested other M3OD models, such as MonoGround and MonoDETR, which do not require extra data.
As shown, our model significantly outperforms the others on the custom dataset, demonstrating the superior generalizability of foundation models.
Notably, MonoGround performs very poorly in 3D object detection, likely due to its assumption that the ground is flat.
In contrast, our model makes no such assumptions, allowing it to achieve the best performance across diverse environments, whether on flat ground or high-bank tracks. Visualized results of the custom dataset are shown in Figure \ref{fig:custom_result_visualized}.

\subsection{Ablation Studies}
\textbf{Effect of the Foundation Model.}
To evaluate the impact of the foundation model for depth estimation and object localization, we compared the MonoDETR model with MonoDETR enhanced by our depth feature extraction module.
In other words, we replaced the visual feature extraction module in our models with an alternative ResNet backbone, instead of using the HFFB.

Table \ref{tab:Ablation1_KITTI} presents the results of the first ablation study on the KITTI dataset. The MonoDETR model enhanced with our depth feature extraction module outperforms the original MonoDETR model. This demonstrates that, with the help of the foundation model, it can better estimate the depth value of each pixel in the image, resulting in improved performance in predicting 3D bounding boxes.
Additionally, by comparing this variant model with our model—which relies solely on DINOv2 as the backbone without an additional ResNet backbone—it can be inferred that foundation models are more effective at capturing visual features than ResNet \cite{he2016deep}. This enhanced feature extraction capability translates to improved performance in predicting 3D bounding boxes.

\begin{table}[h]
\caption[Comparison of models with or without DINOv2 + DPT Head result on the KITTI \textit{val.} set]{Comparison of models with or without DINOv2 + DPT Head result on the KITTI \textit{val.} set for the car class.}
\centering
\small
\begin{tabular}{l|ccc}
\toprule
\multirow{2}{*}{Method} & \multicolumn{3}{c}{Val,\ $AP_{3D}$} \\
& Easy & Mod. & Hard \\
\midrule
MonoDETR\cite{zhang2023monodetr} & 24.29 & 17.52 & 15.28 \\
MonoDETR {\small +DINOv2+DPT Head} & \underline{25.44} & \underline{18.69} & \underline{15.57} \\
\textbf{MonoDINO-DETR} & \textbf{26.72} & \textbf{19.19} & \textbf{15.92} \\
\bottomrule
\end{tabular}
\label{tab:Ablation1_KITTI}
\end{table}

\begin{figure}[h]
    \centering
    \includegraphics[width=0.7\linewidth]{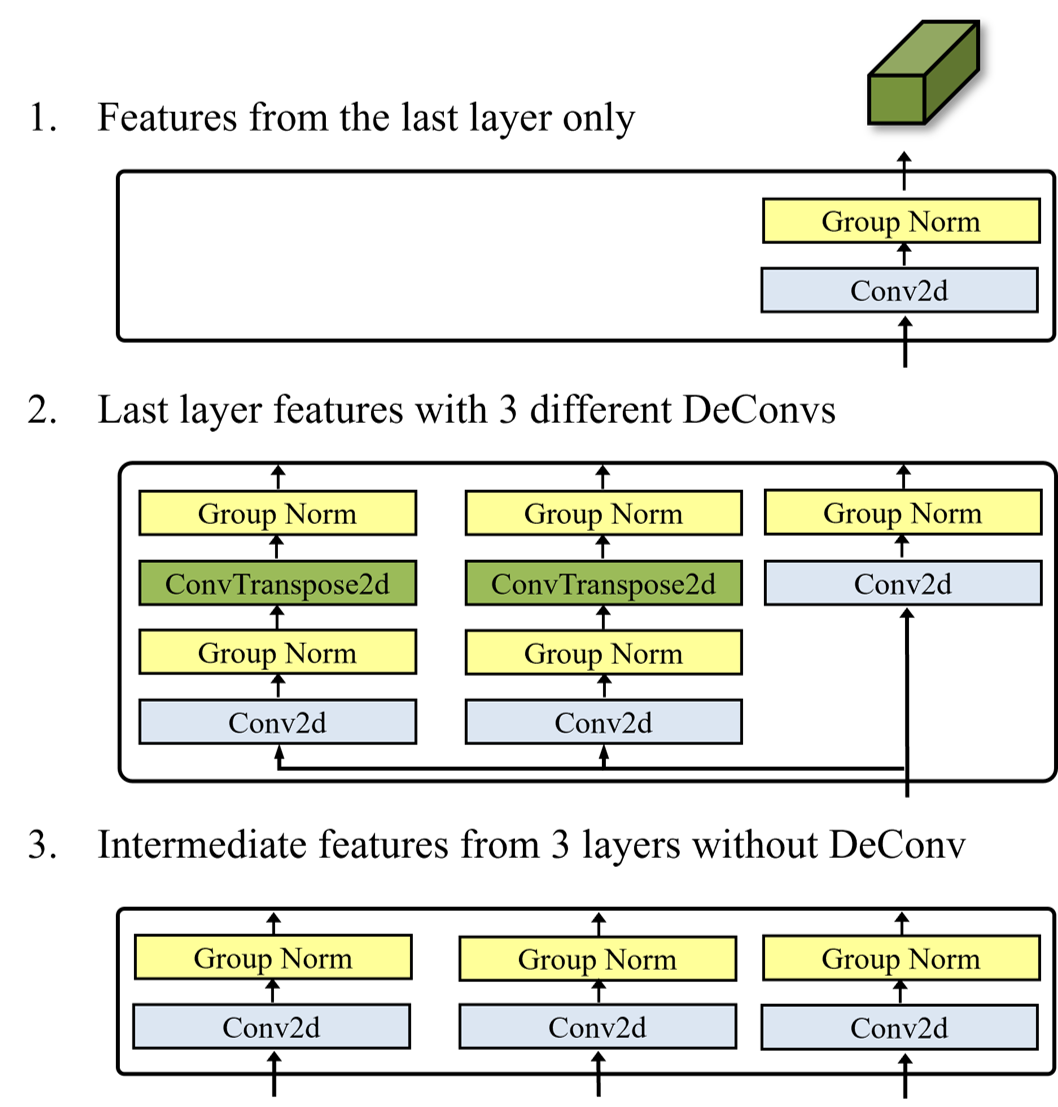}
    \caption[Ablation Study 2: Effect of the HFFB]{\textbf{Ablation Study 2: Effect of the Hierarchical Feature Fusion Block.} We evaluated the effect of HFFB by testing 3 HFFB variants.
    } \label{fig:HFB Ablation}
\end{figure}

\textbf{Effect of the Hierarchical Feature Fusion Block.} To evaluate the impact of the Hierarchical Feature Fusion Block (HFFB), we tested the following variants of HFFB as shown in Figure \ref{fig:HFB Ablation}.

\begin{table*}[h]
\caption[Comparison of models with HFFB variants result on the KITTI \textit{val.} set]{Comparison of models with HFFB variants result on the KITTI \textit{val.} set for the car class.}
\centering
\small
\begin{tabular}{l|ccc|ccc|ccc|c}
	\toprule
\multirow{2}{*}{Method} & \multicolumn{3}{c|}{Val,\ $AP_{3D}$} & \multicolumn{3}{c|}{Val,\ $AP_{BEV}$} & \multicolumn{3}{c|}{Val,\ $AP_{BBOX}$} & Time \\ 
& Easy & Mod. & Hard & Easy & Mod. & Hard & Easy & Mod. & Hard & (ms) \\
\midrule
(1) Last layer only & 23.20 & 16.08 & 12.86 & \underline{34.36} & 23.15 & 19.18 & 92.74 & 77.78 & 70.36 & 49 \\
(2) Last layer with 3 dif. DeConvs & \underline{23.61} & 16.51 & \underline{13.71} & 34.06 & 23.00 & 19.35 & 92.94 & 78.93 & 73.76 & 71 \\
(3) 3 Layers without DeConvs & 21.90 & \underline{16.66} & 13.64 & 33.50 & \underline{23.94} & \underline{20.09} & \underline{95.03} & \underline{83.69} & \underline{76.37} &  51 \\
\midrule
\textbf{3 layers with DeConvs(=HFFB)}  & \textbf{26.72} & \textbf{19.19} & \textbf{15.92} & \textbf{37.65} & \textbf{26.70} & \textbf{21.79} & \textbf{96.00} & \textbf{87.09} & \textbf{79.84} & 66 \\
\textit{Improvement v.s. second-best} & \color{blue}{+3.11} & \color{blue}{+2.53} & \color{blue}{+2.21} & \color{blue}{+3.29} & \color{blue}{+2.76} & \color{blue}{+1.70} & \color{blue}{+0.97} & \color{blue}{+3.40} & \color{blue}{+3.47} & \\
\bottomrule
\end{tabular}
\label{tab:Ablation2_KITTI}
\end{table*}

The first variant uses only the last layer features from the intermediate layers to extract visual features. It does not incorporate any hierarchical architecture to obtain features at diverse resolutions.
The second variant module generates multi-scale features using only the last layer feature.
This architecture resembles a simple feature pyramid of VitDet \cite{li2022exploring}.
The last variant module uses the three layer features from DINOv2 but does not employ transposed convolution to generate different resolutions.

As shown in Table \ref{tab:Ablation2_KITTI}, the HFFB is a key component for localizing objects in 2D space and providing rich visual features to estimate 3D bounding boxes in the decoder module.
Only by combining features from three layers with transposed convolution to generate multi-scale hierarchical features does the model achieve the best performance.
This demonstrates the effectiveness of the HFFB in extracting meaningful local features from DINOv2’s backbone, leading to improved 3D object detection performance.

\textbf{Effect of 6D Dynamic Anchor Boxes.} The final ablation study was conducted to evaluate the effect of 6 dimensional dynamic anchor boxes. We configured the decoder queries as learnable dynamic anchors with six dimensions to accurately estimate bounding boxes. As shown in Table \ref{tab:Ablation3_KITTI}, the model with DAB outperforms the one without it across most metrics and difficulty levels. This demonstrates that 6D DAB effectively incorporates the spatial information of bounding boxes and refines object queries layer by layer, leading to improved performance in 3D object detection.

\begin{table}[h]
\caption[Comparison of models with or without DAB result on the KITTI \textit{val.} set]{Comparison of models with or without DAB result on the KITTI \textit{val.} set for the car class.}
\centering
\small
\begin{tabular}{l|ccc}
\toprule
\multirow{2}{*}{Method} &  \multicolumn{3}{c}{Val,\ $AP_{3D}$} \\
& Easy & Mod. & Hard \\
\midrule
MonoDINO-DETR & 26.72 & 19.19 & 15.92\\
\textbf{MonoDINO-DETR {\small + DAB}} & \textbf{27.93} & \textbf{19.39} & \textbf{15.97} \\
\bottomrule
\end{tabular}
\label{tab:Ablation3_KITTI}
\end{table}

%% file: sections/5.conclusion.tex
\begin{figure*}[h]
    \centering
    \includegraphics[width=14cm]{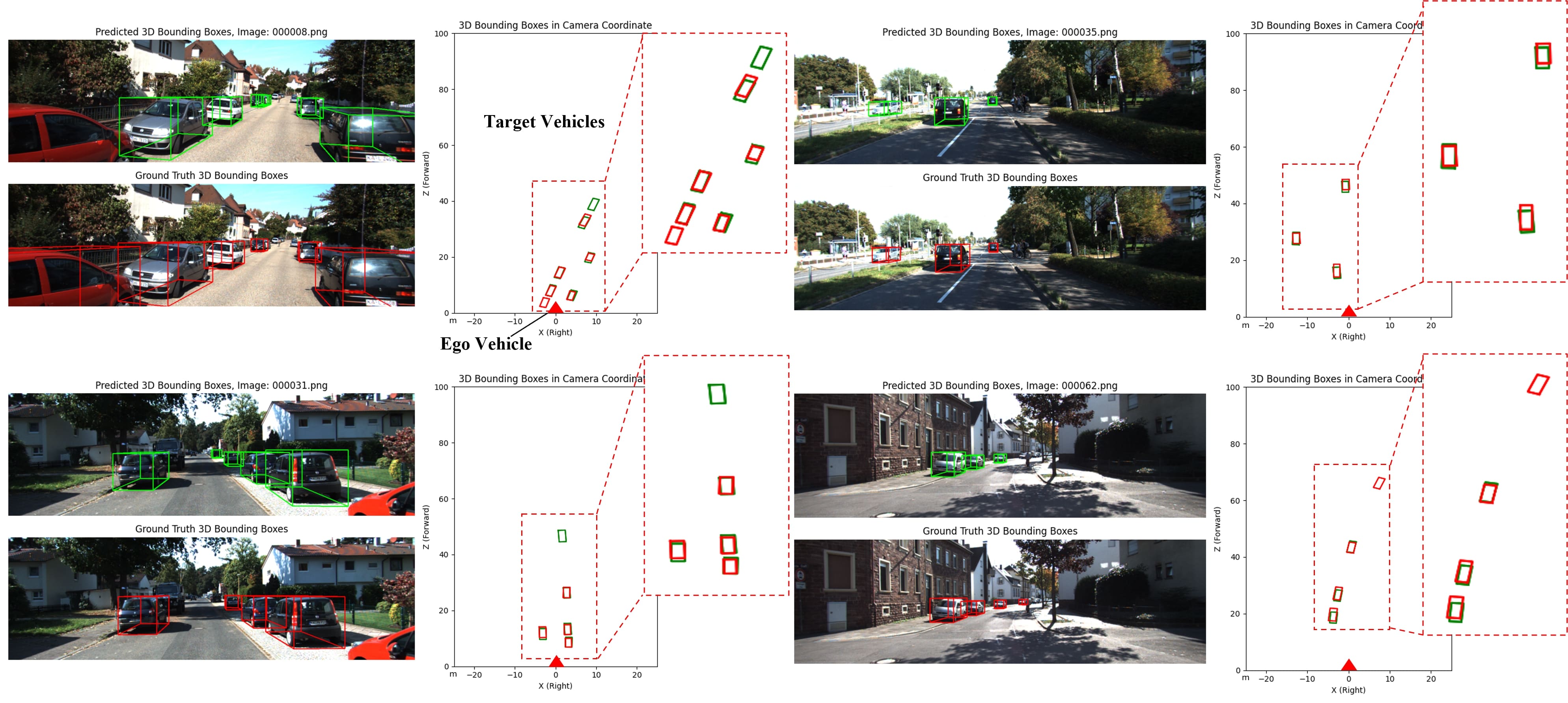}
    \caption[Qualitative Results on KITTI \textit{val.} set]{\textbf{Qualitative results on the KITTI \textit{val} set for the car class.} The proposed method (green) and ground truth (red).
    } \label{fig:KITTI visualized}
\end{figure*}

\begin{figure*}[t]
    \centering
     \includegraphics[width=14cm]{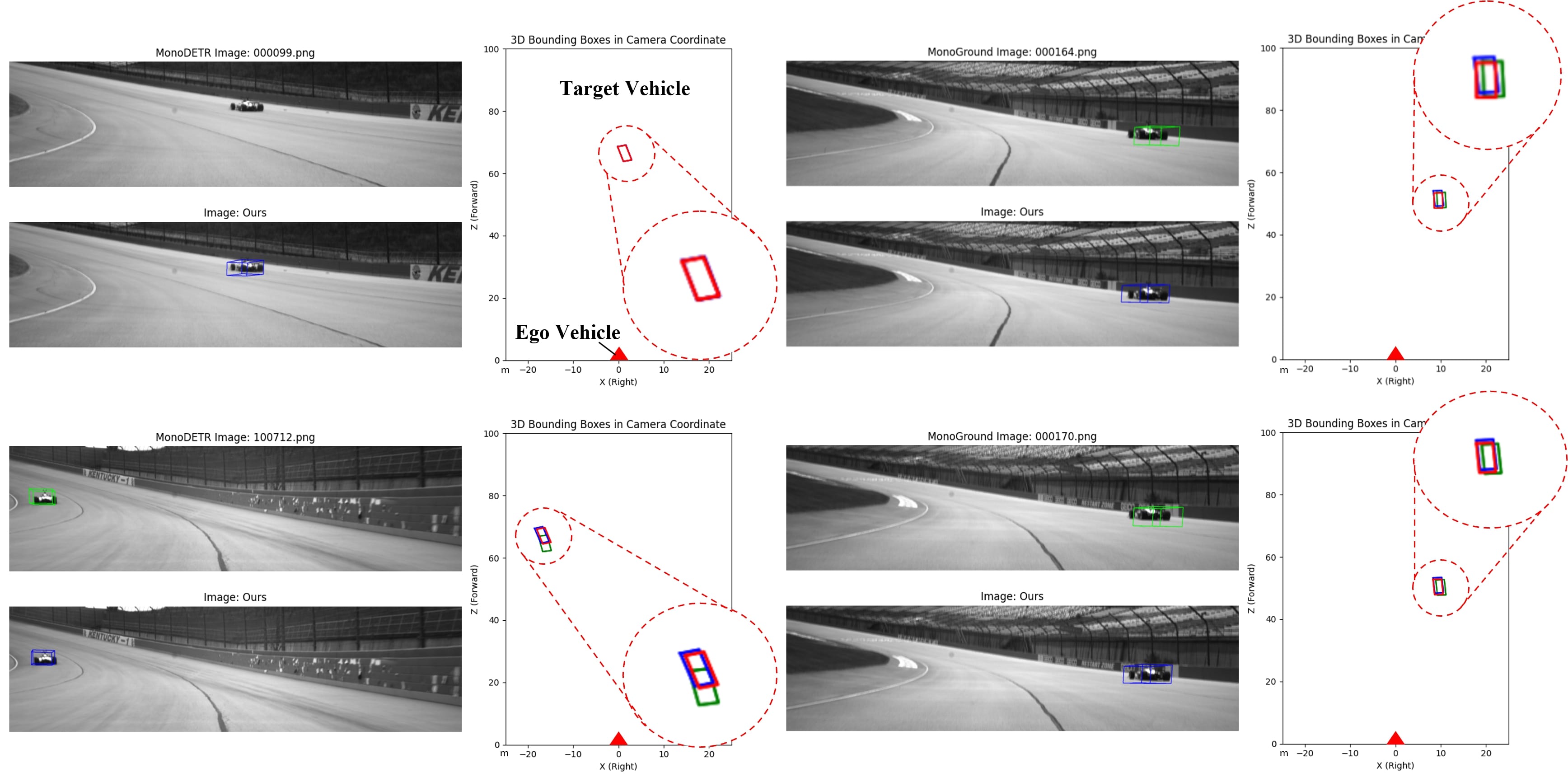}
    \caption{\textbf{Qualitative results on the custom dataset.} Comparison of detection results between the proposed model (blue), the state-of-the-art models (green), and ground truth (red) in ego-view (left) and bird's-eye view (right); MonoDETR (left) and MonoGround (right).}
    \label{fig:custom_result_visualized}
\end{figure*}

\section{Conclusion}
\label{sec:conclusion}
This paper presents a novel approach to monocular 3D object detection by integrating a Vision Foundation Model as the backbone with the DETR architecture, enabling enhanced depth estimation and feature extraction within a single-stage, real-time framework. By incorporating a Hierarchical Feature Fusion Block for multi-scale detection and 6D Dynamic Anchor Boxes for iterative bounding box refinement, the proposed model achieves improved performance without relying on additional data sources, such as LiDAR. Future work will focus on extending the model's capabilities to detect 3D bounding boxes while accounting for rolling and pitching angles.

%% file: root.bbl
\begin{thebibliography}{10}
\providecommand{\url}[1]{#1}
\csname url@samestyle\endcsname
\providecommand{\newblock}{\relax}
\providecommand{\bibinfo}[2]{#2}
\providecommand{\BIBentrySTDinterwordspacing}{\spaceskip=0pt\relax}
\providecommand{\BIBentryALTinterwordstretchfactor}{4}
\providecommand{\BIBentryALTinterwordspacing}{\spaceskip=\fontdimen2\font plus
\BIBentryALTinterwordstretchfactor\fontdimen3\font minus \fontdimen4\font\relax}
\providecommand{\BIBforeignlanguage}[2]{{%
\expandafter\ifx\csname l@#1\endcsname\relax
\typeout{** WARNING: IEEEtran.bst: No hyphenation pattern has been}%
\typeout{** loaded for the language `#1'. Using the pattern for}%
\typeout{** the default language instead.}%
\else
\language=\csname l@#1\endcsname
\fi
#2}}
\providecommand{\BIBdecl}{\relax}
\BIBdecl

\bibitem{carion2020end}
N.~Carion, F.~Massa, G.~Synnaeve, N.~Usunier, A.~Kirillov, and S.~Zagoruyko, ``End-to-end object detection with transformers,'' in \emph{European conference on computer vision}.\hskip 1em plus 0.5em minus 0.4em\relax Springer, 2020, pp. 213--229.

\bibitem{oquab2023dinov2}
M.~Oquab, T.~Darcet, T.~Moutakanni, H.~Vo, M.~Szafraniec, V.~Khalidov, P.~Fernandez, D.~Haziza, F.~Massa, A.~El-Nouby \emph{et~al.}, ``Dinov2: Learning robust visual features without supervision,'' \emph{arXiv preprint arXiv:2304.07193}, 2023.

\bibitem{dosovitskiy2020image}
A.~Dosovitskiy, ``An image is worth 16x16 words: Transformers for image recognition at scale,'' \emph{arXiv preprint arXiv:2010.11929}, 2020.

\bibitem{geiger2012we}
A.~Geiger, P.~Lenz, and R.~Urtasun, ``Are we ready for autonomous driving? the kitti vision benchmark suite,'' in \emph{2012 IEEE conference on computer vision and pattern recognition}.\hskip 1em plus 0.5em minus 0.4em\relax IEEE, 2012, pp. 3354--3361.

\bibitem{brazil2019m3d}
G.~Brazil and X.~Liu, ``M3d-rpn: Monocular 3d region proposal network for object detection,'' in \emph{Proceedings of the IEEE/CVF international conference on computer vision}, 2019, pp. 9287--9296.

\bibitem{qin2022monoground}
Z.~Qin and X.~Li, ``Monoground: Detecting monocular 3d objects from the ground,'' in \emph{Proceedings of the IEEE/CVF Conference on Computer Vision and Pattern Recognition}, 2022, pp. 3793--3802.

\bibitem{ding2020learning}
M.~Ding, Y.~Huo, H.~Yi, Z.~Wang, J.~Shi, Z.~Lu, and P.~Luo, ``Learning depth-guided convolutions for monocular 3d object detection,'' in \emph{Proceedings of the IEEE/CVF Conference on computer vision and pattern recognition workshops}, 2020, pp. 1000--1001.

\bibitem{wang2021depth}
L.~Wang, L.~Du, X.~Ye, Y.~Fu, G.~Guo, X.~Xue, J.~Feng, and L.~Zhang, ``Depth-conditioned dynamic message propagation for monocular 3d object detection,'' in \emph{Proceedings of the IEEE/CVF Conference on Computer Vision and Pattern Recognition}, 2021, pp. 454--463.

\bibitem{huang2022monodtr}
K.-C. Huang, T.-H. Wu, H.-T. Su, and W.~H. Hsu, ``Monodtr: Monocular 3d object detection with depth-aware transformer,'' in \emph{Proceedings of the IEEE/CVF conference on computer vision and pattern recognition}, 2022, pp. 4012--4021.

\bibitem{vaswani2017attention}
A.~Vaswani, ``Attention is all you need,'' \emph{Advances in Neural Information Processing Systems}, 2017.

\bibitem{zhang2023monodetr}
R.~Zhang, H.~Qiu, T.~Wang, Z.~Guo, Z.~Cui, Y.~Qiao, H.~Li, and P.~Gao, ``Monodetr: Depth-guided transformer for monocular 3d object detection,'' in \emph{Proceedings of the IEEE/CVF International Conference on Computer Vision}, 2023, pp. 9155--9166.

\bibitem{reading2021categorical}
C.~Reading, A.~Harakeh, J.~Chae, and S.~L. Waslander, ``Categorical depth distribution network for monocular 3d object detection,'' in \emph{Proceedings of the IEEE/CVF Conference on Computer Vision and Pattern Recognition}, 2021, pp. 8555--8564.

\bibitem{liu2022dab}
S.~Liu, F.~Li, H.~Zhang, X.~Yang, X.~Qi, H.~Su, J.~Zhu, and L.~Zhang, ``Dab-detr: Dynamic anchor boxes are better queries for detr,'' \emph{arXiv preprint arXiv:2201.12329}, 2022.

\bibitem{russakovsky2015imagenet}
O.~Russakovsky, J.~Deng, H.~Su, J.~Krause, S.~Satheesh, S.~Ma, Z.~Huang, A.~Karpathy, A.~Khosla, M.~Bernstein \emph{et~al.}, ``Imagenet large scale visual recognition challenge,'' \emph{International journal of computer vision}, vol. 115, pp. 211--252, 2015.

\bibitem{he2016deep}
K.~He, X.~Zhang, S.~Ren, and J.~Sun, ``Deep residual learning for image recognition,'' in \emph{Proceedings of the IEEE conference on computer vision and pattern recognition}, 2016, pp. 770--778.

\bibitem{huang2017densely}
G.~Huang, Z.~Liu, L.~Van Der~Maaten, and K.~Q. Weinberger, ``Densely connected convolutional networks,'' in \emph{Proceedings of the IEEE conference on computer vision and pattern recognition}, 2017, pp. 4700--4708.

\bibitem{radford2021learning}
A.~Radford, J.~W. Kim, C.~Hallacy, A.~Ramesh, G.~Goh, S.~Agarwal, G.~Sastry, A.~Askell, P.~Mishkin, J.~Clark \emph{et~al.}, ``Learning transferable visual models from natural language supervision,'' in \emph{International conference on machine learning}.\hskip 1em plus 0.5em minus 0.4em\relax PMLR, 2021, pp. 8748--8763.

\bibitem{caron2021emerging}
M.~Caron, H.~Touvron, I.~Misra, H.~J{\'e}gou, J.~Mairal, P.~Bojanowski, and A.~Joulin, ``Emerging properties in self-supervised vision transformers,'' in \emph{Proceedings of the IEEE/CVF international conference on computer vision}, 2021, pp. 9650--9660.

\bibitem{kirillov2023segment}
A.~Kirillov, E.~Mintun, N.~Ravi, H.~Mao, C.~Rolland, L.~Gustafson, T.~Xiao, S.~Whitehead, A.~C. Berg, W.-Y. Lo \emph{et~al.}, ``Segment anything,'' in \emph{Proceedings of the IEEE/CVF International Conference on Computer Vision}, 2023, pp. 4015--4026.

\bibitem{ranftl2021vision}
R.~Ranftl, A.~Bochkovskiy, and V.~Koltun, ``Vision transformers for dense prediction,'' in \emph{Proceedings of the IEEE/CVF international conference on computer vision}, 2021, pp. 12\,179--12\,188.

\bibitem{yang2024depth}
L.~Yang, B.~Kang, Z.~Huang, Z.~Zhao, X.~Xu, J.~Feng, and H.~Zhao, ``Depth anything v2,'' \emph{arXiv preprint arXiv:2406.09414}, 2024.

\bibitem{lin2017refinenet}
G.~Lin, A.~Milan, C.~Shen, and I.~Reid, ``Refinenet: Multi-path refinement networks for high-resolution semantic segmentation,'' in \emph{Proceedings of the IEEE conference on computer vision and pattern recognition}, 2017, pp. 1925--1934.

\bibitem{xian2018monocular}
K.~Xian, C.~Shen, Z.~Cao, H.~Lu, Y.~Xiao, R.~Li, and Z.~Luo, ``Monocular relative depth perception with web stereo data supervision,'' in \emph{Proceedings of the IEEE Conference on Computer Vision and Pattern Recognition}, 2018, pp. 311--320.

\bibitem{lin2017focal}
T.~Lin, ``Focal loss for dense object detection,'' \emph{arXiv preprint arXiv:1708.02002}, 2017.

\bibitem{kuhn1955hungarian}
H.~W. Kuhn, ``The hungarian method for the assignment problem,'' \emph{Naval research logistics quarterly}, vol.~2, no. 1-2, pp. 83--97, 1955.

\bibitem{chen20153d}
X.~Chen, K.~Kundu, Y.~Zhu, A.~G. Berneshawi, H.~Ma, S.~Fidler, and R.~Urtasun, ``3d object proposals for accurate object class detection,'' \emph{Advances in neural information processing systems}, vol.~28, 2015.

\bibitem{chen2016monocular}
X.~Chen, K.~Kundu, Z.~Zhang, H.~Ma, S.~Fidler, and R.~Urtasun, ``Monocular 3d object detection for autonomous driving,'' in \emph{Proceedings of the IEEE conference on computer vision and pattern recognition}, 2016, pp. 2147--2156.

\bibitem{lang2019pointpillars}
A.~H. Lang, S.~Vora, H.~Caesar, L.~Zhou, J.~Yang, and O.~Beijbom, ``Pointpillars: Fast encoders for object detection from point clouds,'' in \emph{Proceedings of the IEEE/CVF conference on computer vision and pattern recognition}, 2019, pp. 12\,697--12\,705.

\bibitem{loshchilov2017decoupled}
I.~Loshchilov, ``Decoupled weight decay regularization,'' \emph{arXiv preprint arXiv:1711.05101}, 2017.

\bibitem{yan2024monocd}
L.~Yan, P.~Yan, S.~Xiong, X.~Xiang, and Y.~Tan, ``Monocd: Monocular 3d object detection with complementary depths,'' in \emph{Proceedings of the IEEE/CVF Conference on Computer Vision and Pattern Recognition}, 2024, pp. 10\,248--10\,257.

\bibitem{li2022exploring}
Y.~Li, H.~Mao, R.~Girshick, and K.~He, ``Exploring plain vision transformer backbones for object detection,'' in \emph{European conference on computer vision}.\hskip 1em plus 0.5em minus 0.4em\relax Springer, 2022, pp. 280--296.

\end{thebibliography}
